\documentclass[letterpaper]{article} % DO NOT CHANGE THIS
\usepackage{aaai2026}  % DO NOT CHANGE THIS
\usepackage{times}  % DO NOT CHANGE THIS
\usepackage{helvet}  % DO NOT CHANGE THIS
\usepackage{courier}  % DO NOT CHANGE THIS
\usepackage[hyphens]{url}  % DO NOT CHANGE THIS
\usepackage{graphicx} % DO NOT CHANGE THIS
\usepackage{natbib}  % DO NOT CHANGE THIS AND DO NOT ADD ANY OPTIONS TO IT
\usepackage{caption} % DO NOT CHANGE THIS AND DO NOT ADD ANY OPTIONS TO IT
\usepackage{algorithm}
\usepackage{algorithmic}

\usepackage{subcaption}
\usepackage{amsmath}
\usepackage{amssymb}
\usepackage{multirow}

\usepackage{algorithm}
\usepackage{algorithmic}
\usepackage{multirow}
\usepackage{multicol}
\usepackage{pifont}
\usepackage{amsfonts,amssymb}
\usepackage{booktabs}

\usepackage{newfloat}
\usepackage{listings}
\DeclareCaptionStyle{ruled}{labelfont=normalfont,labelsep=colon,strut=off} % DO NOT CHANGE THIS
\floatstyle{ruled}
\newfloat{listing}{tb}{lst}{}
\floatname{listing}{Listing}
\title{EGTM: Event-guided Efficient Turbulence Mitigation}
\author{
       Huanan LI\textsuperscript{\rm 1}
       Rui Fan\textsuperscript{\rm 1}, 
       Juntao Guan\textsuperscript{\rm 1,\rm 2},
       Weidong Hao\textsuperscript{\rm 1}, 
       Lai Rui\textsuperscript{\rm 1}\thanks{Corresponding author.},
       Tong Wu\textsuperscript{\rm 1}, 
       Yikai Wang\textsuperscript{\rm 1}, 
       Lin Gu\textsuperscript{\rm 3,\rm 4}$^{*}$,
}
\affiliations{
    \textsuperscript{\rm 1}School of Integrated Circuits, Xidian University, Xi’an 710071, China\\
     \textsuperscript{\rm 2}Hangzhou Institute of Technology, Xidian University, Hangzhou, China\\
    \textsuperscript{\rm 3}RIKEN AIP, Tokyo103-0027, Japan\\
     \textsuperscript{\rm 4}The University of Tokyo, Japan\\
}

\usepackage{bibentry}
\begin{document}
\maketitle

\begin{abstract}
Turbulence mitigation (TM) aims to remove the stochastic distortions and blurs introduced by atmospheric turbulence into frame cameras. Existing state-of-the-art deep-learning TM methods extract turbulence cues from multiple degraded frames to find the so-called ``lucky'', not distorted patch, for ``lucky fusion''. However, it requires high-capacity network to learn from coarse-grained turbulence dynamics between synchronous frames with limited frame-rate, thus fall short in computational and storage efficiency. Event cameras, with microsecond-level temporal resolution, have the potential to fundamentally address this bottleneck with efficient sparse and asynchronous imaging mechanism. In light of this, we (i) present the fundamental \textbf{``event-lucky insight''} to reveal the correlation between turbulence distortions and inverse spatiotemporal distribution of event streams. Then, build upon this insight, we (ii) propose a novel EGTM framework that extracts pixel-level reliable turbulence-free guidance from the explicit but noisy turbulent events for temporal lucky fusion. Moreover, we (iii) build the first turbulence data acquisition system to contribute the first real-world event-driven TM dataset. This demonstrating the great efficiency merit of introducing event modality into TM task. Demo code and data have been uploaded in supplementary material and will be released once accepted.
\end{abstract}

\section{Introduction}
Clear vision is crucial for numerous applications, e.g., urban surveillance~\cite{xu2024long_multi_dataset} and astronomical observation~\cite{xia2025planet}. However, atmospheric turbulence causes highly ill-posed degradations in frame cameras~\cite{gao2025fluidnexus}, manifesting as spatiotemporal ill-posed distortions and random blurs. This presents great challenges for downstream applications, establishing turbulence mitigation (TM) as a pivotal computer vision task.

\begin{figure}[!t]
  \centering
\includegraphics[width=0.99\linewidth]{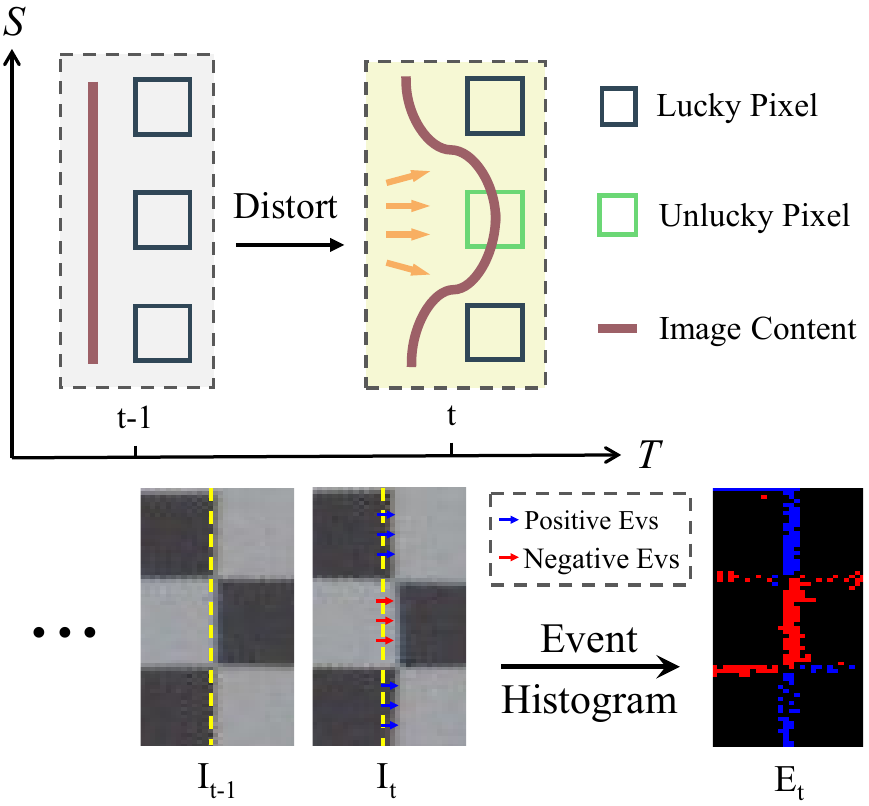}
  \caption{Illustration of turbulent events on a checkerboard image, $I$ is RGB frame and $E$ the event dense visualization.}
\label{fig:2}
\end{figure}

Existing state-or-the-art (SOTA) deep-learning TM solutions~\cite{zhang2024imaging_multi,saha2024turb_sim6_multi,xu2024long_multi_dataset} have demonstrated satisfactory results by exploiting temporal ``lucky fusion'', which identifies potentially turbulence-free (``lucky'') patches or pixels (Figure~\ref{fig:2} top) and fuses them into clean frames. However, they requires heavyweight networks for two reasons: (i) the reliable turbulence cues (i.e., turbulence distortions) essential for plausible TM are latent in the inter-frame changes, thereby requiring high-capacity model to separate them with the other unchanged image contents (Figure~\ref{fig:2} bottom); (ii) the fine-grained, inter-frame turbulence cues can not be obtained due to the limited frame rate, thereby requiring high-capacity model to handle this insufficient temporal diversity~\cite{xia2024nb_sTM4}. This restricts their practical efficiency for real-world deployment.  

\begin{figure}[!t]
    \centering
    \begin{subfigure}{\linewidth}
        \centering
\includegraphics[width=0.99\textwidth]{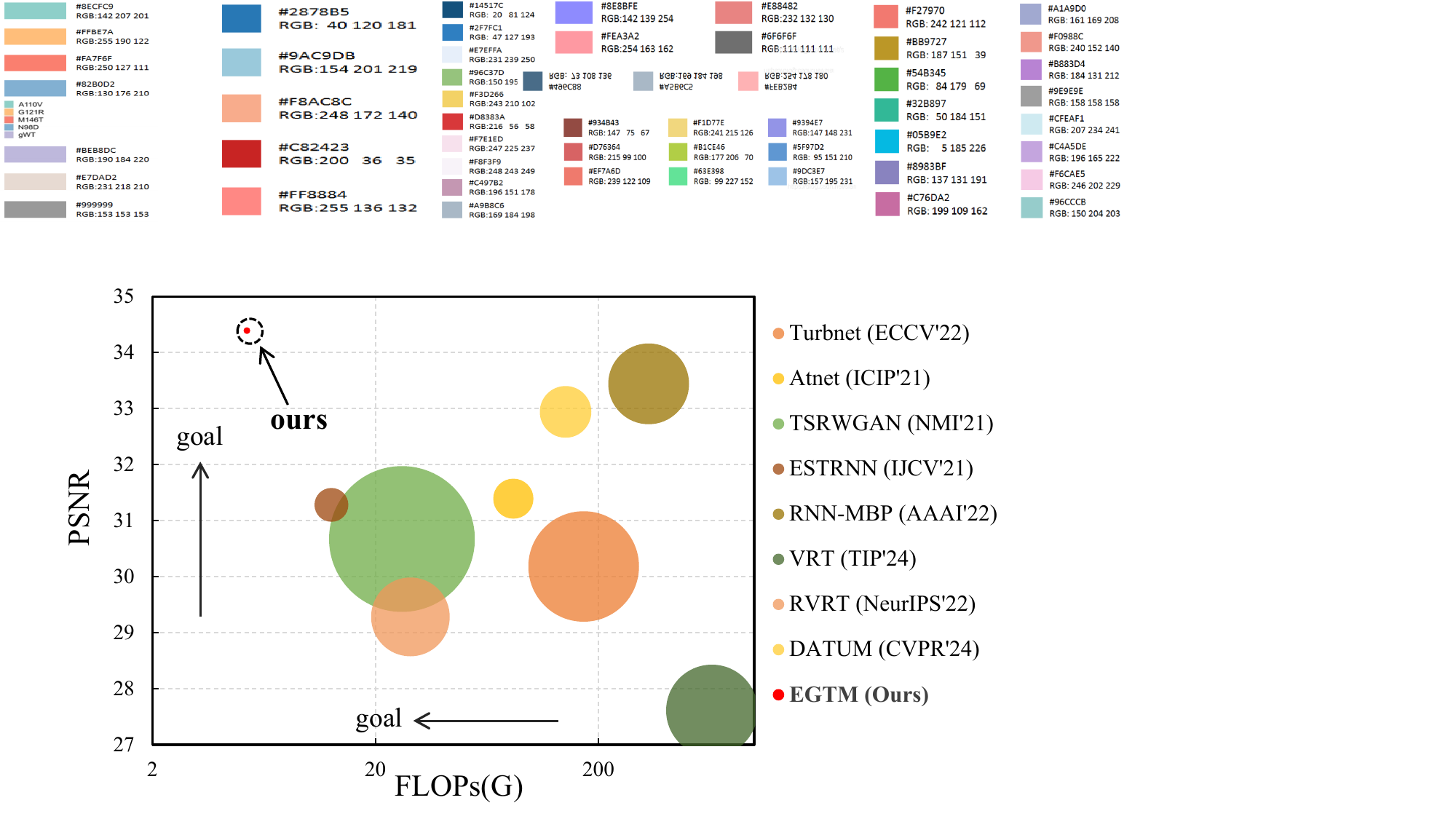}
        \label{fig:1_2}
    \end{subfigure}
    
    \caption{Illustration (top) of our EGTM framework and comparisons on the collected real-world dataset (bottom).}
    \label{fig:1}
\end{figure}

Unlike frame cameras, event cameras asynchronously respond for pixel intensity changes with high temporal resolution~\cite{gallego2020event_survey}, thus can directly and only capture the fine-grained ($~\sim1us$) turbulence distortions regardless of other undistorted regions as illustrated in the bottom of Figure~\ref{fig:2}, thus promising for addressing the efficiency bottleneck of existing frame-based TM methods. Hence, we pioneer the incorporation of event cameras into TM task and build the first \textbf{event-camera based turbulence data acquisition system} (Figure~\ref{fig:3}) to investigate the inherent statistical correlations between the turbulence distortions and event data (Figure~\ref{fig:4}). Then, based on this investigation, we present the fundamental \textbf{``event-lucky insight''}  that the inverse spatiotemporal distribution of turbulent events naturally indicate turbulence-free ``lucky'' regions, which formulate the high correlation of event cameras to turbulence distortions theoretically.

Based on this insight, for static or quasi-static scenes where turbulence dominates the temporal dynamics, we propose an \textbf{event-guided TM framework (EGTM)} (Figure~\ref{fig:5}) to extract reliable turbulence-free guidance from spatiotemporal event distribution for temporal fusion. Due to widely spread noisy events~\cite{duan2024led}, it is challenging to design hand-crafted models to distinguish ``lucky'' regions adaptively. Hence, our framework incorporates a learning-based approach: an Event Distribution Encoding Module (EDEM) transforms raw sparse events into dense representations and extracts pixel-level ``lucky'' guidance through an spatiotemporally separated guidance extraction blocks (SGEB\&TGEB) to re-weight and fuse pixel intensities from multiple turbulent frames. To address residual blur artifacts, we further incorporate a lightweight Details Extraction Block (DEB) that enhance edge details, ensuring comprehensive turbulence mitigation in an end-to-end optimization framework.

To facilitate comprehensive evaluation, we address the current lack of event-based multimodal turbulence datasets by creating both \textbf{simulated and real-world EGTM datasets} through the v2e event simulator~\cite{V2E} and our hybrid event-RGB imaging system. Evaluations on our real-world dataset using NVIDIA RTX 3090 GPU demonstrate that with minimal additional input overhead from sparse event data ($\sim1\%$ compared to RGB frames), our EGTM achieves 2ms inference latency and 0.02M parameters, while showcasing superior restoration quality (+0.94 PSNR, +0.08 SSIM) over SOTA method on our real-world dataset. Our main contributions are:

\begin{itemize}
    \item We reveal the key insight that inverse spatiotemporal distribution of turbulent events indicates turbulence-free cues, enabling highly efficient TM.
    \item We propose the first EGTM framework that extracts reliable lucky fusion weights from event streams for efficient turbulence removal.
    \item We contribute the first synthetic event-driven TM dataset through v2e simulation, and a real-world event-driven TM dataset based on an elaborate turbulence data acquisition system.
    \item Experimental results demonstrate exceptional efficiency gains: 214$\times$ faster inference, 224$\times$ smaller model, and superior restoration quality (+0.94 PSNR, +0.08 SSIM) over existing SOTA multi-frame method.
\end{itemize}

\section{Related Works}
\label{sec:related}

\subsection{Atmospheric Turbulence Mitigation}
\label{sec:turbulence_mitigation}

Atmospheric turbulence mitigation has evolved from classical model-based solutions to modern deep-learning approaches. Early works~\cite{gilles2008atmospheric_model1,oreifej2012simultaneous_model2_moving4,zhu2012removing_model3,anantrasirichai2013atmospheric_model4} exploited the ``lucky effect''~\cite{fried1978probability_luckyeffect}, aggregating sharpest patches through ``lucky fusion''~\cite{aubailly2009automated_luckyfusion}. However, these approaches suffered from limited adaptability and computational inefficiency for complex real-world turbulence.

\begin{figure*}[!t]
  \centering
\includegraphics[width=0.99\textwidth]{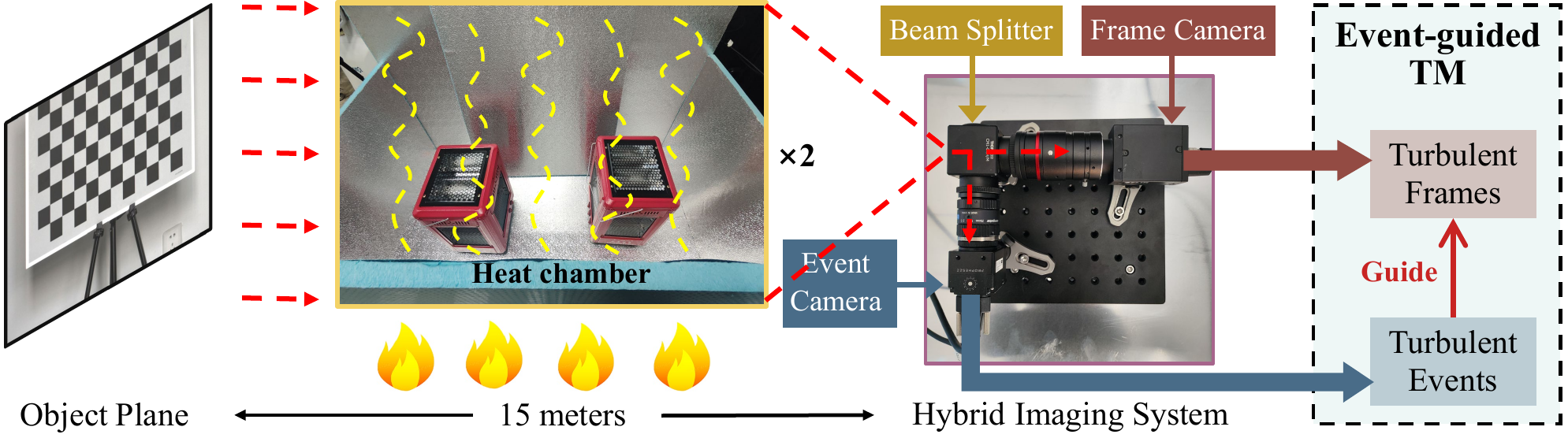}
  \caption{Turbulence data acquisition system and turbulence generation setup. The system integrates event and frame cameras through a beam splitter for synchronized data acquisition, with a heat chamber generating controlled turbulence.}
\label{fig:3}
\end{figure*}

With the support of physics-grounded simulators and large-scale datasets~\cite{saha2024turb_sim6_multi,zhang2024spatio_datum_multi_sim1,jaiswal2023physics_sTM5_sim5}, data-driven solutions significantly outperform classical methods. Current approaches fall into single-frame~\cite{mei2023ltt_sTM1,nair2023ddpm_sTM2,xia2024nb_sTM4,jaiswal2023physics_sTM5_sim5} and multi-frame strategies~\cite{zhang2024imaging_multi,saha2024turb_sim6_multi,xu2024long_multi_dataset,zhang2024spatio_datum_multi_sim1}. Single-frame solutions achieve moderate performance but lack temporal information crucial for complex turbulence patterns, while multi-frame ones demonstrate superior performance by exploiting temporal dynamics. For example, recent method DATUM~\cite{zhang2024spatio_datum_multi_sim1} integrates classical lucky fusion as inductive biases into an end-to-end framework. However, these multi-frame methods fundamentally rely on heavyweight networks to extract implicit turbulence cues from temporally sparse synchronous RGB frames, necessitating complex architectures with substantial computational overhead that limit practical deployment.

\subsection{Event-based Vision for Image Restoration}
\label{sec:event_restoration}

Event cameras, with their asynchronous sensing, high temporal resolution, and low latency~\cite{gallego2020event_survey}, have emerged as powerful tools for various image restoration tasks that analogous to TM. For example, in video deraining~\cite{sun2023event_derain,fu2024event_derain}, events capture precise rain streak trajectories, providing explicit guidance that significantly outperforms frame-based methods. Similarly, event-based deblurring~\cite{zhang2024crosszoom_deblurr,kim2024cmta_deblurr,li2024coarse_deblurr} leverages motion-triggered events for accurate blur kernel estimation. These successes demonstrate a key advantage: event cameras capture explicit spatiotemporal dynamics that remain implicit in frame-based observations, reducing the need for complex implicit feature learning. However, despite proven effectiveness in these related domains, event cameras still remain unexplored for TM tasks, presenting significant opportunities for efficiency improvements of TM.

\subsection{Turbulence Datasets and Multi-modal Challenges}
\label{sec:turbulence_datasets}
Real-world turbulence datasets have advanced robust TM solutions, including heat-generated turbulence~\cite{EFF,vidalmata2020bridging_TurbText,mao2022single_sTM_dataset} and long-range natural turbulence~\cite{mao2020image_model6_moving2_multi,xu2024long_multi_dataset}. However, these datasets exclusively contain frame sequences without corresponding event data, presenting a fundamental challenge for developing event-guided TM frameworks. Hence, it is necessary to create new multimodal datasets with synchronized event-frame data under controlled turbulence conditions, essential for training and evaluating the proposed EGTM framework.

\section{Method}
\label{sec:Method}

%In this section, we detail our hybrid imaging system and dataset collection strategy, then validate the core insight linking event distributions to turbulence-free regions, and finally introduce our EGTM framework.

\subsection{Turbulence Data Acquisition System}
\label{sec:Data Collection}

To enable synchronized event-frame turbulence mitigation research, we developed a event-camera based turbulence data acquisition system, as shown in Figure~\ref{fig:3}. It consist of a heat chamber and hybrid imaging system. 
To provide high-quality and controlled visual inputs for dataset construction, we implemented a screen to display the dataset image. For controlled turbulence generation, we implemented a heat chamber system positioned 15 meters from the target, generating temperature gradients of 20-60°C following established protocols~\cite{Mao2022SingleFA}. Ground-truth images are captured with the heat source disabled, while test scenes are automatically displayed via computer-controlled screens to ensure consistent experimental conditions.
In terms of the hybrid imaging system, we designed a co-optical axis imaging system comprising an event camera (Prophesee EVK5), frame camera (FLIR Grasshopper3), and beam splitter (Thorlabs BSW26R).
By dividing the incoming light into two equal portions, the beam splitter sends one portion to the event camera and the remaining portion to the conventional camera respectively.
The system achieves spatial alignment through region-of-interest cropping followed by stereo rectification, while temporal synchronization utilizes programmable trigger circuits with microsecond precision ($< 10\mu s$). 

\begin{figure}[!t]
  \centering
\includegraphics[width=0.99\linewidth]{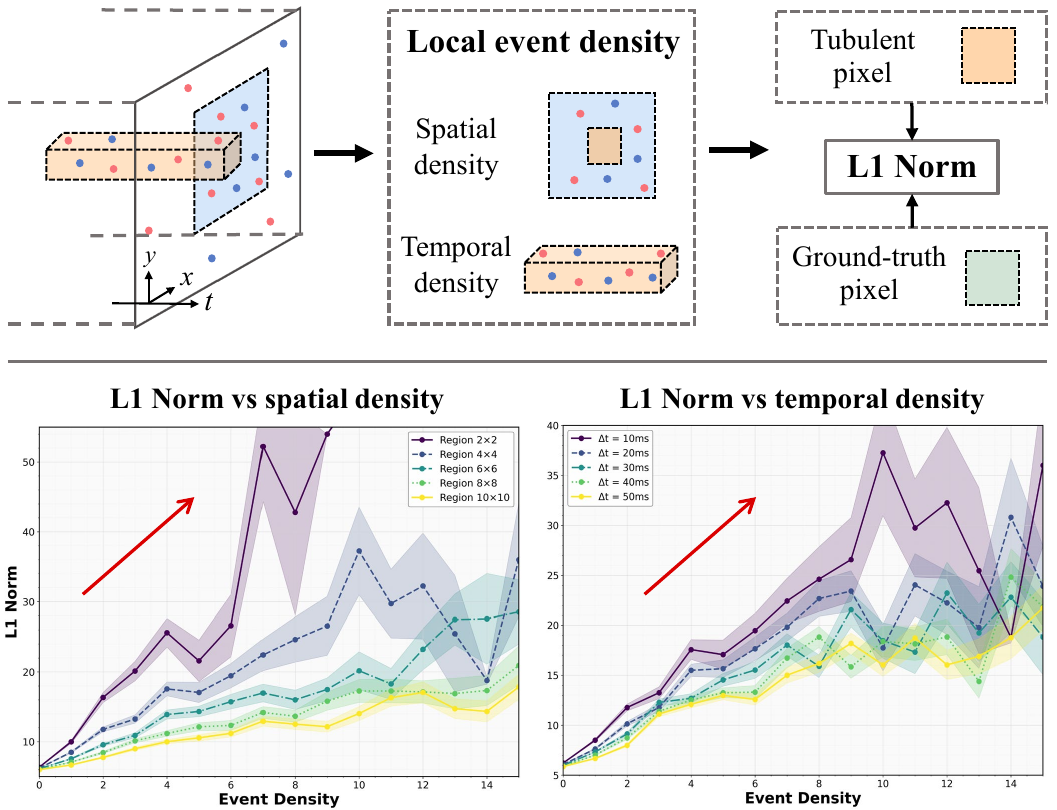}
  \caption{Empirical validation of inverse correlation between event density and turbulence degradation.}
\label{fig:4}
\end{figure}

\begin{figure*}[!t]
  \centering
\includegraphics[width=0.99\textwidth]{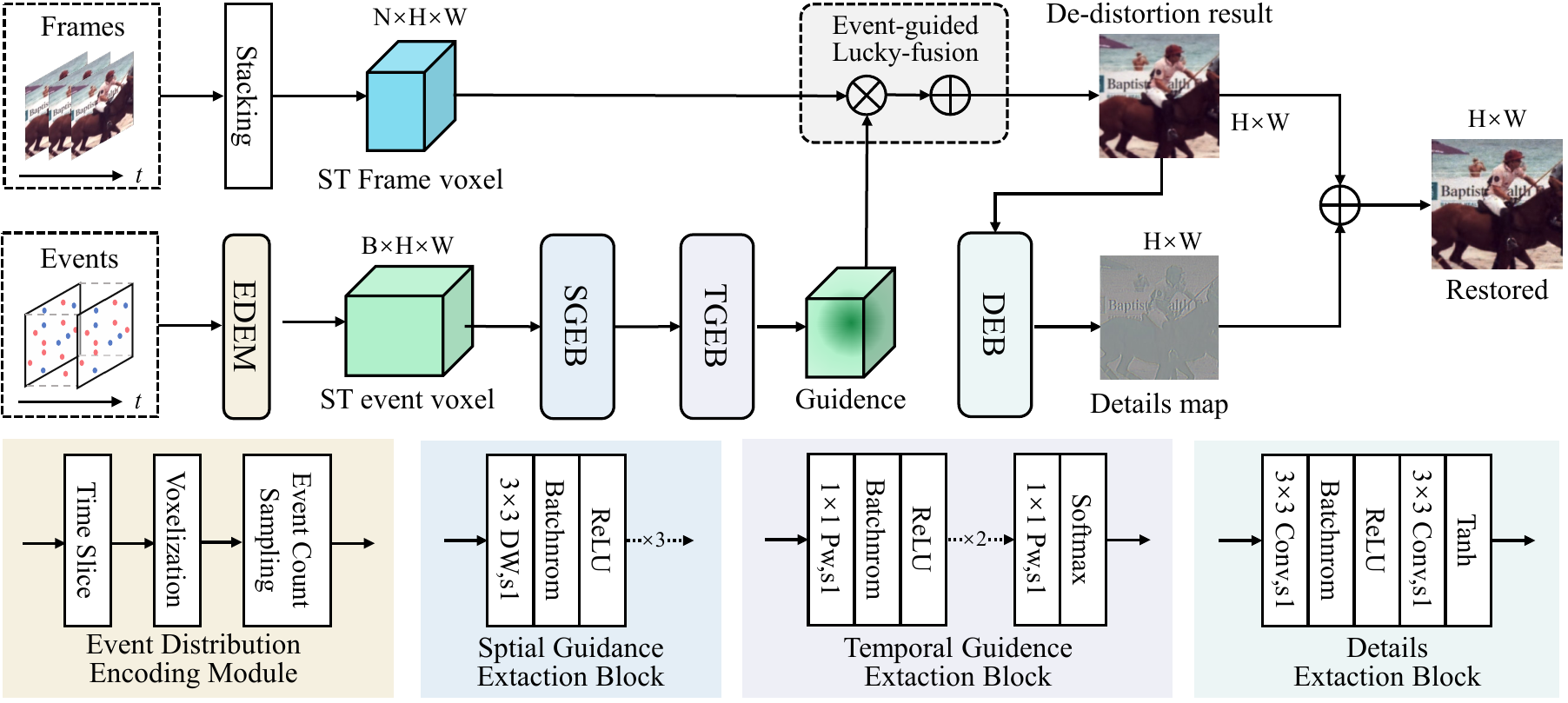}
  \caption{Architecture of our EGTM. It comprises EDEM for converting sparse events to dense voxel representations, dual guidance blocks (SGEB/TGEB) for extracting spatial and temporal lucky region weights, and DEB for details refinement.}
\label{fig:5}
\end{figure*}

The resulting real-world dataset contains 793 samples from MIT Places~\cite{zhou2017places}, each comprising 36 turbulent frames (20 FPS, 1.8s duration), synchronized event streams, and a ground-truth reference image cropped at $256\times256$ pixels size. Data quality is validated through spatial-temporal consistency checks and manual inspection. The dataset is divided into 634/159 samples for training/testing with scene-level splitting to prevent overfitting. Additionally, we generate a synthetic benchmark by applying the v2e simulator~\cite{V2E} to the ATSyn-static dataset~\cite{zhang2024spatio_datum_multi_sim1} for comprehensive evaluation. These two complementary benchmarks provide a solid foundation for event-driven turbulence mitigation research.

\subsection{``Lucky Insight'' Meets Event Cameras}
\label{sec:``Lucky Insight'' Meets Event Cameras}

The stochastic nature of turbulence creates ``lucky'' regions where image patches remain relatively undistorted~\cite{zhang2024imaging_multi}. Traditional multi-frame approaches struggle to identify these regions efficiently due to limited temporal resolution and implicit turbulence dynamics~\cite{aubailly2009automated_luckyfusion,zhang2024imaging_multi}. Event cameras offer a unique advantage: in static scenes, triggered events primarily result from turbulence-induced brightness changes, as shown in Figure~\ref{fig:2}. This leads to our core hypothesis: \textbf{regions with fewer events are more likely to be turbulence-free ``lucky''}.

To validate this hypothesis, we conducted correlation analysis on 100 randomly selected samples from our real-world dataset (sufficient for detecting medium effect sizes with 80\% power), each containing 500 uniformly sampled pixels from ground-truth and turbulent frames with corresponding events. As illustrated in Figure~\ref{fig:4}, we computed the L1 norm between turbulent and ground-truth pixels, correlating reconstruction error with local event density in spatiotemporal regions. We examined spatial event density across regions from 2$\times$2 to 10$\times$10 within 100ms temporal windows, and temporal event density across time windows ($\Delta t$ = 10ms to 50ms) for fixed 1$\times$1 spatial regions. Both analyses demonstrate significant positive correlation between event density and reconstruction error (Pearson correlation coefficients ranging from 0.52 to 0.78 under statistical significance $p < 0.001$), confirming that regions with lower event density are more likely to be turbulence-free ``lucky'' regions. Despite variability due to turbulence stochasticity and event noise, the consistent positive trend provides empirical foundation for our event-guided framework.

Interestingly, we also observe that larger spatial regions (10$\times$10) achieve more stable correlation ($r = 0.68$) due to improved statistical robustness, while shorter temporal windows ($~\sim$10ms) show higher correlation ($r = 0.61$) by capturing fine-grained turbulence dynamics. However, overly large spatiotemporal regions reduce correlation strength due to information dilution. This analysis reveals that effective turbulence mitigation requires adaptive spatiotemporal reasoning that can balance local perception with statistical robustness-a capability that motivates our CNN-based EGTM framework design detailed in the next section.

\begin{table*}[t]
\scriptsize
\centering
    \begin{tabular}{l|c|ccc|cc|cc}
    \hline
        \multirow{2}{*}{\textbf{Method}} & \multirow{2}{*}{\textbf{Input Type}} & \multicolumn{3}{c|}{\textbf{Efficiency}} & \multicolumn{2}{c|}{\textbf{Synthetic}} & \multicolumn{2}{c}{\textbf{Real-world}} \\
        & & \#Params(M)↓ & GFLOPs↓ & Latency(ms)↓ & PSNR↑ & SSIM↑ & PSNR↑ & SSIM↑ \\
        \hline
        TurbNet(ECCV'22)& Single-frame & 26.6 (1330$\times$) & 171.9 (115$\times$) & 48 (24$\times$) & 20.13 & 0.5145 & 30.18 & 0.8839 \\
        ATnet(ICIP'21)& Single-frame & 3.5 (175$\times$) & 83.1 (55$\times$) & 23 (11$\times$) & 19.66 & 0.5559 & 31.39 & 0.9043 \\
        TSRWGAN(NMI'21)& Multi-frame & 46.3 (2315$\times$) & 26.3 (17$\times$) & 33 (16$\times$) & 23.16 & 0.8407 & 31.90 & 0.9169 \\
        VRT(TIP'24)& Multi-frame & 18.3 (915$\times$) & 646.1 (431$\times$) & 2813 (1406$\times$) & 24.27 & 0.8641 & 31.61 & 0.9302 \\
        RNN-MBP(AAAI'22)& Multi-frame & 14.2 (710$\times$) & 336.3 (224$\times$) & 428 (214$\times$) & 24.64 & 0.8775 & \underline{33.44} & \underline{0.9262} \\
        ESTRNN(IJCV'21)& Multi-frame & 2.5 (125$\times$) & 12.7 (8.4$\times$) & 50 (25$\times$) & 26.23 & 0.9017 & 31.28 & 0.9095 \\
        RVRT(NeurIPS'22)& Multi-frame & 13.6 (680$\times$) & 93.7 (62$\times$) & 291 (145$\times$) & 25.71 & 0.8876 & 29.28 & 0.8926 \\
        DATUM(CVPR'24)& Multi-frame & 5.8 (290$\times$) & 142.5 (95$\times$) & 97 (48$\times$) & \underline{26.76} & \underline{0.9102} & 33.17 & 0.9215 \\
        \textbf{EGTM(Ours)} & \textbf{Multi-frame\&Events} & \textbf{0.02 (1.0$\times$)} & \textbf{1.5 (1.0$\times$)} & \textbf{2 (1.0$\times$)} & \textbf{26.88} & \textbf{0.9152} & \textbf{34.38} & \textbf{0.9339} \\
        \hline
    \end{tabular}
    \caption{Quantitative comparisons on synthetic and real-world datasets. Best results are in \textbf{bold} and second-best are \underline{underlined}.}
    \label{tab:quant}
    \end{table*}

\subsection{Event-guided TM Framework}
\label{sec:EGTM Framework}

As illustrated in Figure~\ref{fig:5}, based on the validated event-turbulence correlation, our EGTM comprises three key components: Event Distribution Encoding Module (EDEM) for converting sparse events to dense representations, dual guidance blocks (SGEB/TGEB) for extracting spatial and temporal lucky region weights, and Details Extraction Block (DEB) for residual artifact refinement.

\textbf{Event Distribution Encoding Module.} Raw event streams are inherently sparse, requiring dense representation for dense neural processing~\cite{fan2025eventpillars}. EDEM aims to transform sparse events into structured spatiotemporal voxels through time-slice voxelization. Given event stream $\mathcal{E} = \{(x_i, y_i, t_i, p_i)\}$ where $(x_i, y_i)$ denotes spatial coordinates, $t_i$ timestamp, and $p_i$ polarity, we partition the temporal dimension into $B$ uniform bins with bin length $\Delta t = T_{total}/B$, where $T_{total}$ spans the entire sequence duration. Each event is assigned to bin $b = \lfloor t_i / \Delta t \rfloor$ and accumulated spatially to generate event density voxel $\mathbf{V} \in \mathbb{R}^{B \times H \times W}$:
\begin{equation}
\mathbf{V}[b, y, x] = \sum_{e_{j}} \delta(b-b_j,y-y_j,x-x_j),e_{j}\in \mathcal{E},
\end{equation}

This dense representation of event distribution effectively capture temporal dynamics while maintaining spatial locality, enabling subsequent CNNs processing.

\textbf{Spatial Guidance Extraction Block.} SGEB extracts spatial distributional features by processing event voxels through grouped depth-wise separable convolutions to maintain computational efficiency. The design motivation stems from the need to capture both local spatial patterns within individual bins. This architecture enables the network to firstly identify spatial regions with consistently low event density across multiple temporal bins, effectively highlighting turbulence-free areas:
\begin{equation}
\mathbf{F}_s = \text{SGEB}(\mathbf{V})
\end{equation}
where $\mathbf{F}_s\in \mathbb{R}^{B \times H \times W}$ represents extracted spatial features encoding regional event distribution patterns. The detailed architecture is shown in Figure~\ref{fig:5}.

\textbf{Temporal Guidance Extraction Block.} As shown in Figure~\ref{fig:5}, TGEB generates pixel-level temporal fusion weights by progressively reducing the temporal dimension through $1 \times 1$ convolutions with channel reduction. This design enables efficient temporal aggregation while preserving spatial resolution. The progressive channel reduction ($B \rightarrow B/2 \rightarrow B/4 \rightarrow N$) allows the network to gradually compress temporal information while maintaining discriminative power. The final softmax normalization ensures that fusion weights sum to unity across the temporal dimension, providing probabilistic interpretation of frame reliability:
\begin{equation}
\mathbf{W} = \text{TGEB}(\mathbf{F}_s)
\end{equation}
where $\mathbf{W} \in \mathbb{R}^{N \times H \times W}$ is the temporal fusion weights.

\textbf{Event-guided Lucky Fusion.} The extracted guidance weights $\mathbf{W}$ are applied to the input frame sequence $\mathbf{I} = \{\mathbf{I}_1, \mathbf{I}_2, \ldots, \mathbf{I}_N\}$ for weighted temporal fusion:
\begin{equation}
\mathbf{I}_{fused} = \sum_{i=1}^{N} \mathbf{W}_i \odot \mathbf{I}_i
\end{equation}
where $\odot$ denotes element-wise multiplication and $\mathbf{W}_i\in \mathbb{R}^{H \times W}$ represents the $i$-th temporal weight map, emphasizing pixels with lower spatiotemporal event density.

\textbf{Details Extraction Block.} While event-guided fusion effectively removes primary turbulence distortions, residual blur artifacts may persist due to the temporal averaging effect~\cite{zhang2024spatio_datum_multi_sim1}. DEB employs a lightweight residual refinement network to enhance fine details and texture information. As shown in Figure~\ref{fig:5}, the block uses shallow convolutions with tanh activation to learn residual corrections, ensuring that the refinement process preserves the main structural integrity while enhancing details:
\begin{equation}
\mathbf{I}_{final} = \mathbf{I}_{fused} + \text{DEB}(\mathbf{I}_{fused})
\end{equation}

\begin{figure*}[!t]
  \centering
\includegraphics[width=0.99\linewidth]{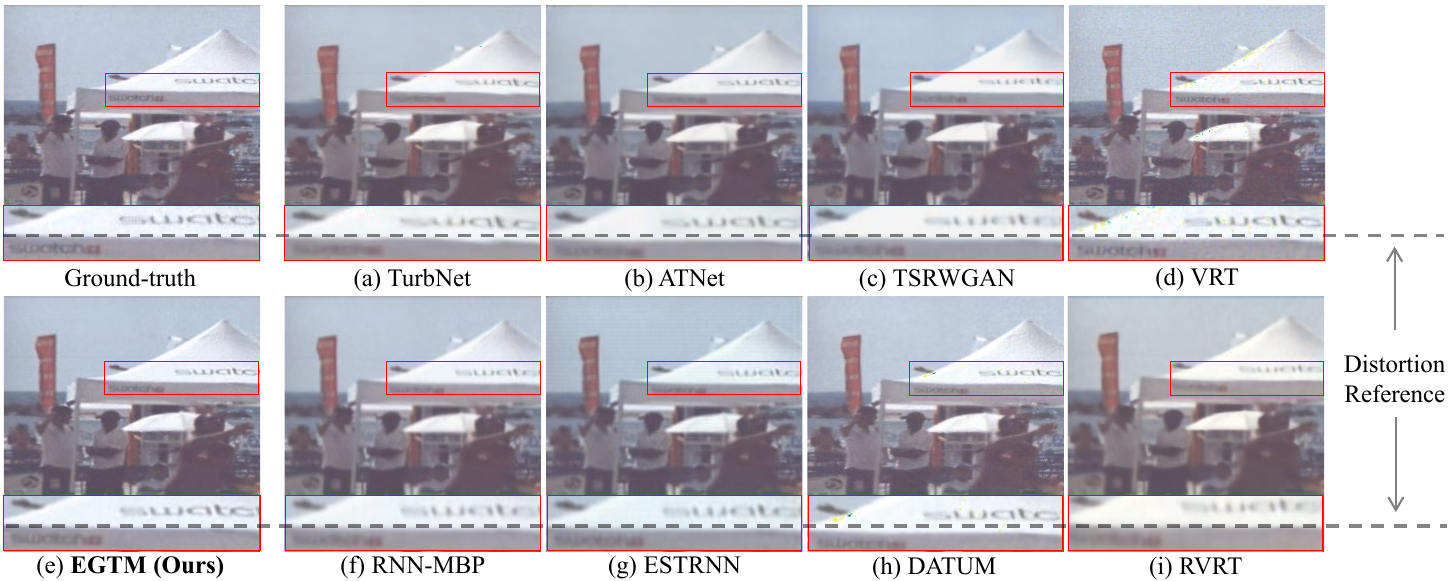}
  \caption{Qualitative comparison on our \textbf{real-world} event-frame turbulence dataset.}
\label{fig:6}
\end{figure*}

\textbf{Loss Function.} Our end-to-end EGTM framework is optimized using a combination of reconstruction loss and perceptual loss:
\begin{equation}
\mathcal{L} = \mathcal{L}_{\text{rec}} + \lambda \mathcal{L}_{\text{perc}}
\end{equation}
where $\mathcal{L}_{\text{rec}} = \|\mathbf{I}_{final} - \mathbf{I}_{gt}\|_1$ ensures pixel-level fidelity, $\mathcal{L}_{\text{perc}}$ based on VGG features preserves perceptual quality, and $\lambda = 0.3$ balances the two terms.

\section{Experiments}
\label{sec:experiments}

\subsection{Experimental Setup}
\label{sec:setup}

\subsubsection{Network Architecture} 
As illustrated in Figure~\ref{fig:5}, the EDEM transforms raw event streams into dense spatiotemporal voxels with $B=4N$ bins, where $N=11$ represents both the number of input frames and the temporal window size for training and evaluation. The SGEB maintains consistent $B \rightarrow B \rightarrow B$ channel dimensions through three spatial encoding layers for spatial guidance extraction. The TGEB performs progressive channel reduction from $B \rightarrow B/2 \rightarrow B/4 \rightarrow N$ to generate pixel-wise lucky fusion weights with softmax normalization. The DEB contains a lightweight detail enhancer with $3 \rightarrow 16 \rightarrow 16 \rightarrow 3$ channel progression for final output refinement.

\subsubsection{Training \& Evaluation}
All compared methods are trained from scratch using their original network architectures on our datasets to ensure fair comparison. We implement them using their publicly available code with original hyperparameters adapted to our 11-frame input setting. Our EGTM framework is trained using the Adam optimizer with $\beta_1=0.9$ and $\beta_2=0.999$. The initial learning rate is set to $5 \times 10^{-3}$ with cosine annealing scheduling. Training is conducted for 20 epochs with batch size 32 on PyTorch using an NVIDIA RTX 3090 GPU. Both training and evaluation consistently use 11 frames as input to maintain temporal consistency across all experimental conditions. All the experimental results are conducted and averaged on 5 different random seeds to mitigate the robustness and reproducibility of our framework.

\begin{table}[t]
\centering
\begin{tabular}{l|cc}
\hline
Components & PSNR↑ & SSIM↑ \\
\hline
Inverse Voxel & 26.73 & 0.7451 \\
+SGEB & 28.52 & 0.7914 \\
+TGEB & 29.76 & 0.8038 \\
+SGEB\&TGEB & 32.25 & 0.9203 \\
All Components & \textbf{34.38} & \textbf{0.9339} \\
\hline
\end{tabular}
\caption{Component ablation study on real-world dataset.}
\label{tab:ablation_components}
\end{table}

\subsection{Comparison with State-of-the-Art Methods}
\label{sec:comparison}

\subsubsection{Quantitative Comparison}
Table~\ref{tab:quant} presents comprehensive quantitative comparisons on both synthetic and real-world datasets, evaluating restoration quality (PSNR \& SSIM) and computational efficiency (parameters \& FLOPs). The results demonstrate that multi-frame methods consistently outperform single-frame approaches (TurbNet, ATnet), validating the importance of temporal information for TM. Our EGTM achieves comparable performance on synthetic data (+0.12 PSNR, +0.005 SSIM vs. DATUM) while significantly surpassing all baseline methods on real-world datasets (+0.94 PSNR, +0.08 SSIM vs. RNN-MBP) with statistical significance $p<0.05$. This performance gap suggests that real-world turbulent events provide more reliable fine-grained turbulence cues that synthetic events cannot fully capture.

Regarding efficiency, EGTM achieves exceptional performance with only 0.02M parameters and 1.5G FLOPs, representing a 125× reduction in model size and 8.4× improvement in computational efficiency compared to the lightest method ESTRNN~\cite{zhong2023real}. This superiority stems from our event-guided design that extracts explicit turbulence cues, eliminating the need for heavyweight architectures required by implicit learning approaches. The sparse nature of event data contributes minimal input overhead ($\sim$1\% compared to frames) while providing crucial guidance, making our method both effective and practical.

\subsubsection{Qualitative Comparison}
Figure~\ref{fig:6} presents visual comparisons on representative real-world examples captured under our hybrid imaging system. The results demonstrate that our EGTM achieves superior turbulence mitigation performance in practical conditions. Specifically, methods like TurbNet~\cite{mao2022single_sTM_dataset}, ATnet~\cite{nair2021confidence_multi} and RVRT~\cite{liang2022recurrent} exhibit both significant distortion and blurring, while TSRWGAN~\cite{jin2021neutralizing_multi}, RNN-MBP~\cite{zhu2022deep} and ESTRNN~\cite{zhong2023real} show reduced distortion but persistent blur. DATUM~\cite{zhang2024spatio_datum_multi_sim1} and VRT~\cite{liang2024vrt} achieve sharper results but suffer from overall spatial misalignment (distortion) and introduce additional noise and artifacts. In contrast, our EGTM produces results closest to ground truth with minimal distortion and blur, demonstrating the effectiveness of our approach for identifying and leveraging turbulence-free regions.

\subsection{Ablation Studies}
\label{sec:ablation}

\subsubsection{Component Module Analysis}
Table~\ref{tab:ablation_components} presents comprehensive ablation studies evaluating each component's contribution on the real-world dataset. Results demonstrate that each module contributes significantly to overall performance. Removing event guidance (using inverse raw event voxel as weights) severely degrades performance (26.73 PSNR), confirming the critical role of our learning-based guidance extraction. The SGEB provides essential spatial reasoning capabilities (+1.79 PSNR), while TGEB enables temporal weight generation (+1.24 PSNR). The combination of SGEB and TGEB achieves synergistic improvements (+5.52 PSNR), and the DEB provides crucial final refinement (+2.13 PSNR). These results highlight the necessity of our neural framework design for extracting reliable lucky fusion weights from noisy turbulent events.

\subsubsection{Lucky Weight Visualization}
\begin{figure}[!t]
  \centering
\includegraphics[width=0.99\linewidth]{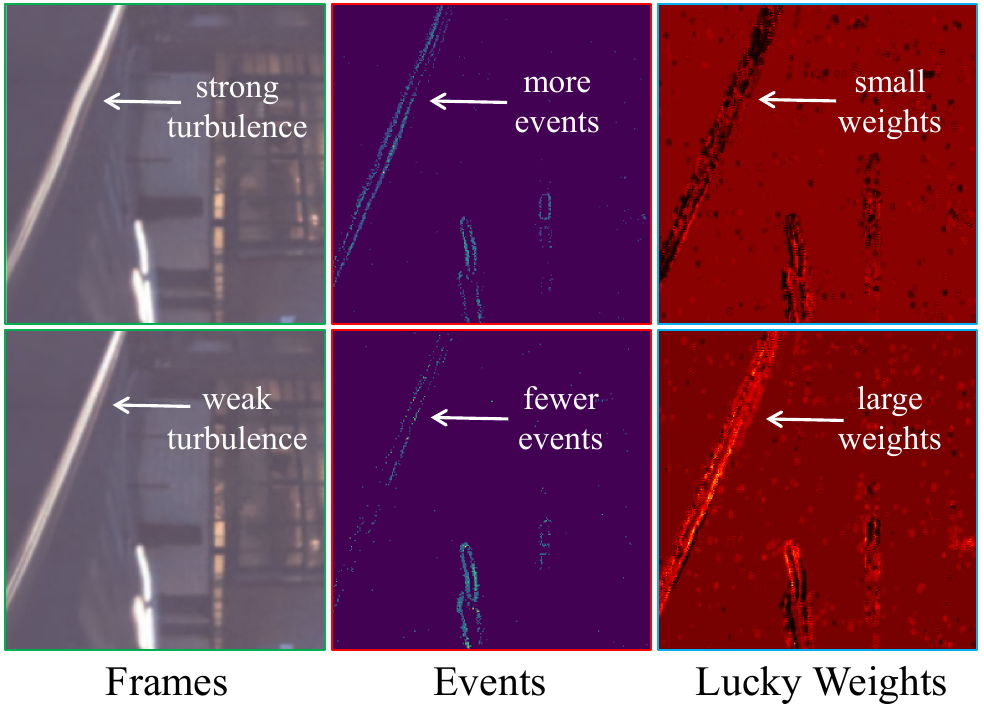}
  \caption{Visualization of the learned lucky weights through EGTM and the corresponding turbulent event distribution.}
\label{fig:9}
\end{figure}

Figure~\ref{fig:9} provides visualizations of our event-guided processing pipeline, validating the effectiveness of lucky weight extraction. The original event distribution shows higher density in turbulent regions with frequent intensity fluctuations. Our EGTM successfully extracts lucky weights that inversely correlate with event density, effectively identifying stable regions. The visualization confirms that our method automatically identifies and leverages ``lucky'' regions with minimal distortion, providing a principled approach to turbulence mitigation.

\section{Conclusion}
\label{sec:conclu}

This paper presents EGTM, the first event-guided framework for atmospheric turbulence mitigation, built upon the key insight that sparse event distributions inversely correlate with turbulence-free ``lucky'' regions. Our framework combines Event Distribution Encoding Module (EDEM) with dual guidance blocks (SGEB/TGEB) to extract pixel-level lucky region weights, achieving superior performance and efficiency compared to multi-frame approaches. The accompanying EGTM dataset provides the first synchronized event-frame turbulence benchmark for the community. Beyond turbulence mitigation, this work establishes a new paradigm demonstrating how event cameras' unique temporal dynamics can address fundamental limitations in traditional vision systems, opening promising directions for multimodal computational imaging.

\bibliography{Reference.bib}
\end{document}